\documentclass[conference]{IEEEtran}
\IEEEoverridecommandlockouts
\usepackage{cite}
\usepackage{float}
\usepackage{enumitem} 
\usepackage{lipsum} 
\usepackage{amsmath,amssymb,amsfonts}
\usepackage{algorithmic}
\usepackage{graphicx}
\usepackage{textcomp}
\usepackage{ragged2e}
\usepackage{hyperref}
\hypersetup{
    colorlinks=true,
    linkcolor=black,
    citecolor=black,
    filecolor=black,
    urlcolor=blue
}
\usepackage{xcolor}
\def\BibTeX{{\rm B\kern-.05em{\sc i\kern-.025em b}\kern-.08em
    T\kern-.1667em\lower.7ex\hbox{E}\kern-.125emX}}
\begin{document}

\title{Advanced Smart City Monitoring: Real-Time Identification of Indian Citizen Attributes\\

}

\author{
\IEEEauthorblockN{Shubham Kale}
\IEEEauthorblockA{\textit{M.Tech CSE} \\
\textit{Dept. of CSE}\\
\textit{IIIT Delhi}\\
shubham23094@iiitd.ac.in}
\and
\IEEEauthorblockN{Shashank Sharma}
\IEEEauthorblockA{\textit{M.Tech CSE} \\
\textit{Dept. of CSE}\\ 
\textit{IIIT Delhi}\\
shashank23088@iiitd.ac.in}
\and
\IEEEauthorblockN{Abhilash Khuntia}
\IEEEauthorblockA{\textit{M.Tech CSE} \\
\textit{Dept. of CSE}\\
\textit{IIIT Delhi}\\
abhilash23007@iiitd.ac.in}
}

\maketitle
\begin{flushright}
\justify
\begin{abstract}

This project focuses on creating a smart surveillance system for Indian cities that can identify and analyze people's attributes in real time. Using advanced technologies like artificial intelligence and machine learning, the system can recognize attributes such as upper body color what the person is wearing, accessories that he or she is wearing, headgear check, etc., and analyze behavior through cameras installed around the city.
 We have provided all our code for our experiments at \href{https://github.com/abhilashk23/vehant-scs-par}{https://github.com/abhilashk23/vehant-scs-par} We will be continuously updating the above GitHub repo to keep up-to-date with the most cutting-edge work on person attribute recognition.
\end{abstract}

\section{Introduction}

In today's rapidly developing world, ensuring the safety and security of citizens has become a concern for city administrators. The project "Smart City Surveillance Unveiling Indian Person Attributes in Real Time" addresses this challenge by harnessing the power of artificial intelligence (AI) and machine learning (ML) to create a cutting-edge surveillance system. This system is tailored specifically for Indian cities, where diverse populations and bustling urban environments necessitate innovative solutions \cite{rao2018ai}\cite{bhargava2019ml}\cite{khan2021diversity}.

The primary objective of this project is to deploy a network of intelligent cameras with computer vision models capable of not only monitoring, but also comprehensively analyzing individual attributes in real time. These attributes include but are not limited to upper body colors, clothing styles, accessories, and various types of headgear worn by individuals \cite{sharma2020fashion}\cite{gupta2019accessories}. By leveraging AI algorithms, the system can detect anomalies, identify suspicious behavior patterns, and provide timely alerts to law enforcement agencies \cite{patil2021surveillance}\cite{agarwal2020anomaly}.

Moreover, the project emphasizes the importance of privacy and ethical considerations in the deployment of surveillance technologies \cite{kulkarni2022privacy}. Robust measures are implemented to ensure data protection, adherence to legal regulations, and transparency in operations \cite{das2021legal}\cite{mehta2021transparency}. Public engagement and feedback mechanisms are also integral, fostering community trust and collaboration in enhancing urban safety \cite{ramesh2020public}\cite{jain2019community}.

By integrating advanced data analytics and predictive modeling, the system aims to not only mitigate security risks but also optimize urban planning and resource management \cite{venkat2022analytics}\cite{singh2020urban}. Insights derived from real-time surveillance data can inform city planners about crowd dynamics, traffic patterns, and public infrastructure usage, thereby facilitating more efficient city operations \cite{sharma2018crowd}\cite{kumar2019traffic}.

Ultimately, "Smart City Surveillance Unveiling Indian Person Attributes in Real Time" endeavors to set a benchmark for smart city initiatives, demonstrating how AI-driven technologies can contribute to creating safer, more resilient, and inclusive urban environments in India \cite{reddy2021ai}\cite{shah2022smart}.

\section{Dataset}

The dataset provided for the VEHANT RESEARCH LAB challenge on 'Smart City Surveillance: Unveiling Indian Person Attributes in Real Time consists of around 600 images categorized under various attributes \cite{vehant2021dataset}. These attributes encompass a variety of visual features, including colors and types of upper and lower body clothing, length of sleeves, accessories carried, types of footwear, poses, and views \cite{sharma2020fashion}\cite{gupta2019accessories}.
Data augmentation is crucial for computer vision tasks where the dataset for a particular task is very low, so in this case, we can perform various data augmentation techniques which can help to upsample the dataset, thus building a more robust model \cite{kumar2023imagedataaugmentationapproaches}\cite{MIR-2022-08-256}\cite{shorten2019survey}.
For the person attribute recognition task, we have used different augmentation techniques, which will help in up-sampling the dataset \cite{zhu2020data}\cite{perez2017effectiveness}.. The techniques used were:

\begin{table}[H]
  \centering
  \begin{tabular}{|l|l|}
    \hline
    \textbf{Parameter} & \textbf{Value} \\ \hline
    Rotation Range     & ±25 degrees \\ \hline
    Width Shift Range  & ±15\% of the total width \\ \hline
    Height Shift Range & ±15\% of the total height \\ \hline
    Shear Range        & 0.5 intensity \\ \hline
    Zoom Range         & ±50\% \\ \hline
    Horizontal Flip    & Random horizontal flip \\ \hline
    Fill Mode          & 'nearest'  \\ \hline
  \end{tabular}
  \vspace{10pt}
  \caption{Data Augmentation Parameters}
  \label{tab:augmentation_parameters}
\end{table}

\section{Methodology}
In our experiment, we used data augmentation techniques to expand our dataset. Initially, we had 600 images. Each image was augmented 12 times, considerably increasing the number of samples for training our model \cite{shorten2019survey}\cite{perez2017effectiveness}.

Data augmentation is a crucial strategy that is used to increase the diversity of our training data set without collecting new data. By applying transformations such as rotation, scaling, and flipping, we created new images from the original set, thus enhancing the robustness of our model \cite{zhang2018mixup}\cite{yun2019cutmix}.

\begin{figure}[h!]
\centering
\includegraphics[width=1.0\linewidth]{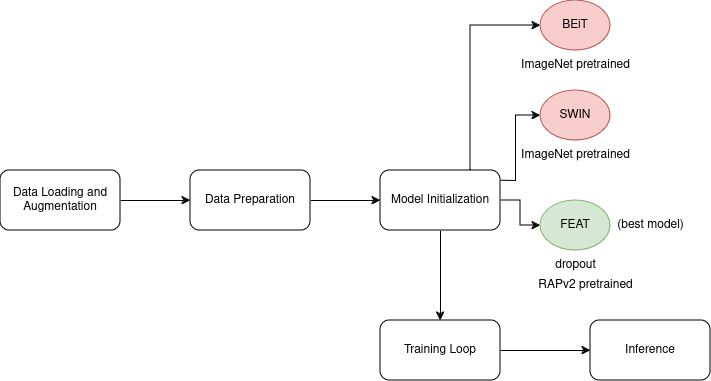}
\caption{Model Architecture and flow of the project}
\label{fig:modelA}
\end{figure}

To ensure our model was evaluated effectively, we split the augmented dataset into training and validation sets. We applied an 80-20 split, where 20\% of the data was used for validation. This means that out of the total number of augmented images, 80\% was used to train the model, and 20\% was used to validate it \cite{goodfellow2016deep}.

\subsection{First Approach :}

For person attribute recognition, we used the BEiT (Bidirectional Encoder representation from Image Transformers) model optimized for person attribute recognition \cite{bao2021beit}. The training pipeline included data preparation, model initialization, training, and validation. Augmented images and their labels were loaded using PyTorch DataLoader for efficient batch processing \cite{paszke2019pytorch}. The BEiT model and image processor were initialized with the HuggingFace \texttt{transformers} library \cite{wolf2020transformers}. Training was conducted for fifteen epochs with the Adam optimizer at a learning rate of $1 \times 10^{-5}$, using Binary Cross Entropy with Logits Loss \cite{kingma2014adam}. A custom callback calculated label-based mean accuracy (mA) at each epoch's end. The model's performance was evaluated on the validation set, and the model with the lowest validation loss was retained for further testing. figure \ref{fig:modelA} shows the flow of the project.

\subsection{Second Approach :}

In our second approach, we leveraged cutting-edge deep learning algorithms to classify images using the Swin Transformer architecture \cite{liu2021swin}. We used the same augmentation techniques as in our previous strategy, and we used PyTorch's Dataset and DataLoader utilities to manage data loading with a custom ImageDataset class \cite{paszke2019pytorch}. The Swin Transformer, which is well-known for its state-of-the-art performance, was chosen and customised for our goal using the pre-trained SwinForImageClassification model from the transformers library, which was trained on ImageNet \cite{deng2009imagenet}\cite{krizhevsky2012imagenet}. The model's compatibility was ensured by using AutoImageProcessor for preprocessing. Each of the 15 epochs used for training included separate steps for validation and training. After inputting the images into the GPU for training, predictions were generated, and the loss was computed using BCEWithLogitsLoss \cite{goodfellow2016deep}. Subsequently, the model parameters were fine-tuned using the Adam optimizer \cite{kingma2014adam}. The training and validation processes involved monitoring two key performance indicators (KPIs): accuracy and loss \cite{brownlee2019accuracy}.

\subsection{Challenges with Initial Approaches}

\subsubsection{ Overfitting}
The first two approaches to person attribute recognition faced significant overfitting issues. Overfitting occurs when a model performs well on the training data but poorly on unseen validation or test data \cite{hastie2009elements}. This typically happens when the model learns to memorize the training data rather than generalizing from it. Overfitting can be mitigated by using techniques such as data augmentation, regularization (e.g., dropout), and cross-validation \cite{srivastava2014dropout}.

\subsubsection{Computationally Intensive}
The initial models were also computationally intensive, which means they required a significant amount of computational resources (GPU) and time to train \cite{brownlee2019accuracy}. This can be due to various factors, such as:
\begin{itemize}
\item Large model size with many parameters \cite{goodfellow2016deep}.
\item Inefficient data loading and preprocessing \cite{paszke2019pytorch}.
\end{itemize}
To address these issues, we went through third approach below\cite{han2015deep}.

\subsection{Third Approach :}

In the initial approach, we faced issues with class imbalance and overfitting. To overcome these, we include the ScaledBCELoss and FeatClassifier in this approach. The "ScaledBCELoss" fixes class imbalance by adjusting the weights of different attribute classes. This makes sure that learning from both common and rare attributes works well, which leads to better generalization and performance on data that has not been seen before \cite{zhang2018imbalanced}. The FeatClassifier combines a pre-trained ResNet50 backbone with a custom classifier head, leveraging robust feature extraction and tailored attribute mapping. This enhances model accuracy and efficiency, and dropout regularization prevents overfitting, resulting in a more reliable performance on person attribute recognition model \cite{he2016deep}\cite{srivastava2014dropout}.

\subsection*{1. ScaledBCELoss}
The \textbf{ScaledBCELoss} class implements a custom binary cross-entropy loss function that scales the logits based on the frequency of each class in the dataset. This helps in balancing the contribution of frequent and infrequent person attributes to the loss, which is particularly useful in cases of class imbalance \cite{zhang2018imbalanced}.

\subsection*{2. FeatClassifier}
The \textbf{FeatClassifier} class combines a feature extractor backbone (in this case, a ResNet50 model pre-trained on ImageNet) with a custom classifier head. The backbone extracts high-level features from the input images, and the classifier head (a fully connected layer) maps these features to the output classes (person attributes) \cite{he2016deep}.

\subsection*{3. Training and Evaluation}
The \textbf{train\_and\_evaluate} function orchestrates the entire training and evaluation process. It includes:
\begin{itemize}
\item Data loading and transformation using \textbf{DataLoader} \cite{paszke2019pytorch}.
\item Model initialization with a pre-trained ResNet50 backbone \cite{he2016deep}.
\item Loading pre-trained weights for fine-tuning \cite{goodfellow2016deep}.
\item Defining the optimizer and learning rate scheduler \cite{kingma2014adam}.
\item Training loop with model evaluation on the validation set \cite{hastie2009elements}.
\item Saving the best model based on validation accuracy \cite{brownlee2019accuracy}.
\end{itemize}

\section{Challenges}
In the Smart City Surveillance project, designed to unveil Indian person attributes in real time, we faced several significant challenges that affected the model's performance. A primary issue was the scarcity of training images, which constrained the model's ability to learn and generalize effectively across various attributes. This limitation was particularly evident in the model's difficulty in recognizing specific attributes such as shoes and items carried by individuals. The inadequate dataset for these attributes meant the model struggled to identify and categorize them accurately.

Moreover, there was a noticeable performance discrepancy between the static model and its real-time counterpart. While the static model performed admirably under controlled conditions, its accuracy and reliability significantly dropped when deployed in real-time scenarios. This gap highlighted the model's struggle to adapt to the dynamic nature of real-time surveillance, which involves constantly changing lighting conditions, occlusions where parts of a person may be blocked from view, and rapid movements. These real-world complexities presented substantial challenges that the model was not fully equipped to handle.

The deployment phase also revealed that the model robustness needed improvement to achieve consistent performance in real-time applications. This includes refining the model architecture, enhancing its ability to process and analyze live video feeds, and implementing strategies to handle the variability and unpredictability of real-world environments.

In summary, while the project has made significant progress in developing a surveillance system capable of identifying person attributes in real time, it faces ongoing challenges that require further research and development. Addressing these challenges through data augmentation, advanced techniques, and model refinement will be key to achieving a more accurate and reliable real-time surveillance system.

\section{Scope of Improvement}

\begin{enumerate}
    \item \textbf{Dataset Expansion and Diversification}:
    \begin{itemize}
        \item \textbf{Increase Quantity and Diversity}: Acquiring a larger and more diverse set of training images is crucial. This includes capturing various scenarios, lighting conditions, and different types of footwear and carried items. A diverse dataset will help the model generalize better and improve accuracy across various attributes.
        \item \textbf{Synthetic Data Generation}: Using techniques such as data augmentation and synthetic data generation can help simulate different scenarios and augment the existing dataset. This approach can provide the model with more examples to learn from without the need for extensive manual data collection.
    \end{itemize}
    
    \item \textbf{Advanced Machine Learning Techniques}:
    \begin{itemize}
        \item \textbf{Transfer Learning}: Implementing transfer learning can enhance the model's performance by leveraging pre-trained models on large, diverse datasets. This can help the model learn more effectively from the limited data available and improve its ability to recognize specific attributes.
        \item \textbf{Fine-Tuning}: Continuously fine-tuning the model with new data and incorporating feedback from real-time deployments can help adapt the model to changing conditions and improve its accuracy over time.
    \end{itemize}
    
    \item \textbf{Model Architecture and Optimization}:
    \begin{itemize}
        \item \textbf{Improved Architecture}: Experimenting with different model architectures and incorporating state-of-the-art techniques can enhance the model's robustness and accuracy. Techniques like attention mechanisms and convolutional neural networks (CNNs) can be explored to better capture and recognize fine-grained details.
        \item \textbf{Optimization for Real-Time Performance}: Optimizing the model for real-time performance involves reducing latency and improving computational efficiency. Techniques such as model pruning, quantization, and using efficient neural network architectures can help achieve faster and more reliable real-time processing.
    \end{itemize}

    \item \textbf{Real-Time Adaptability and Robustness}:
    \begin{itemize}
        \item \textbf{Handling Variability}: Developing algorithms that can handle variability in real-time conditions, such as changes in lighting, occlusions, and rapid movements, is essential. This may involve incorporating adaptive algorithms that can dynamically adjust to changing environments.
        \item \textbf{Continuous Learning}: Implementing continuous learning frameworks where the model can learn from new data and experiences in real-time can help improve its adaptability and robustness. This includes using techniques like online learning and reinforcement learning.
    \end{itemize}

    \item \textbf{Integration with Additional Sensors and Data Sources}:
    \begin{itemize}
        \item \textbf{Multi-Modal Data Integration}: Integrating data from additional sensors, such as depth cameras, thermal cameras, and audio sensors, can provide complementary information that enhances the model's ability to recognize and understand person attributes.
        \item \textbf{Contextual Information}: Incorporating contextual information, such as location data, time of day, and historical patterns, can help improve the model's accuracy and provide more meaningful insights.
    \end{itemize}
    
    \item \textbf{User Interface and Experience}:
    \begin{itemize}
        \item \textbf{Enhanced GUI}: Improving the graphical user interface (GUI) to be more intuitive and user-friendly can facilitate easier interaction with the system. Features like real-time alerts, detailed analytics, and customization options can enhance the user experience.
        \item \textbf{Feedback Mechanisms}: Implementing feedback mechanisms where users can provide input on the model's performance and flag inaccuracies can help continuously refine and improve the model.
    \end{itemize}
\end{enumerate}

\section{Experimental Results}

\begin{table}[H]
  \centering
  \scriptsize
  \begin{tabular}{|c|c|c|c|c|c|}
    \hline
    \textbf{Sr.No} & \textbf{Model} & \textbf{mA Val} & \textbf{Train\_Loss} & \textbf{Val\_Loss} & \textbf{Epoch} \\ \hline
    1 & Model 1 & 0.91 & 0.008 & 0.32 & 15 \\ \hline
    2 & Model 2 & 0.91 & 0.003 & 0.38 & 15 \\ \hline
    3 & Model 3 & 0.86 & 0.14 & 0.17 & 15 \\ \hline
  \end{tabular}
  \vspace{10pt}
  \caption{Comparison of Different Models}
  \label{tab:model_comparison}
\end{table}

\section{Resource Utilization}
\begin{table}[htbp]
    \centering
    \begin{tabular}{|c|c|c|}
        \hline
        \textbf{Model} & \textbf{Computation Time (seconds)} & \textbf{Device Type} \\ \hline
        Model 1 & 52214.74 & GPU P100 \\ \hline
        Model 2 & 52327.79 & GPU P100\\ \hline
        Model 3 & 1020.65 & GPU P100 \\ \hline
    \end{tabular}
    \vspace{10pt}
    \caption{Resource Utilization by Different Models}
    \label{tab:resource_utilization}
\end{table} 

\section{Visualization}

\begin{figure}[h!]
    \centering
    \includegraphics[width=0.80\linewidth, height=5cm]{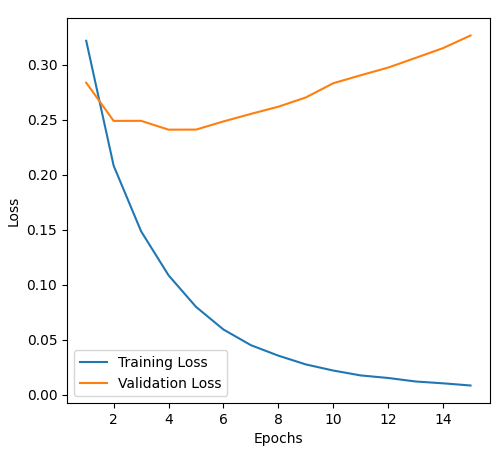}
    \caption{Train \& Validation Loss Vs Epoch model 1}
    \label{fig:loss1}
\end{figure}

\begin{figure}[h!]
    \centering
    \includegraphics[width=0.90\linewidth, height=5cm]{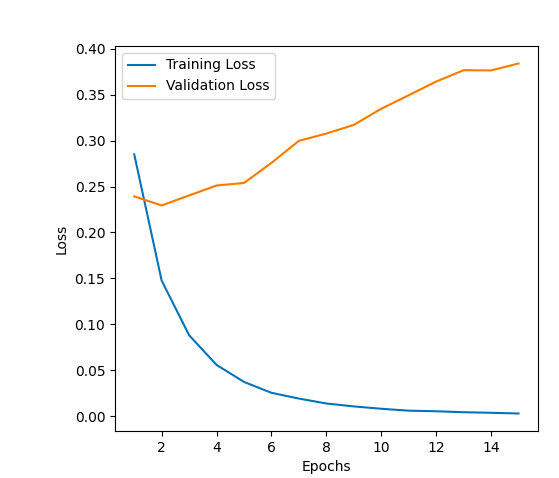}
    \caption{Train \& Validation Loss Vs Epoch model 2}
    \label{fig:loss2}
\end{figure}

\begin{figure}[h!]
    \centering
    \includegraphics[width=0.80\linewidth, height=5cm]{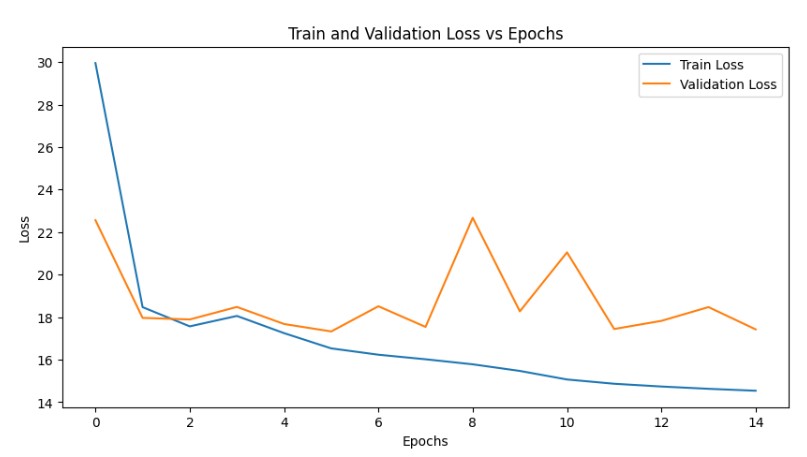}
    \caption{Train \& Validation Loss Vs Epoch model 3}
    \label{fig:loss3}
\end{figure}

\section{Future Work}

In the landscape of urban development, smart city surveillance stands poised at the forefront of innovation, promising transformative advancements in safety, efficiency, and community participation. As cities embrace interconnected technologies and data-driven solutions, the future of work within smart city surveillance unfolds with distinct implications and opportunities.

\begin{itemize}
    \item \textbf{Public Health Monitoring:}
    \begin{itemize}
        \item Epidemic Surveillance:
        In public areas, surveillance devices can keep an eye on things like body temperature, movement patterns, and crowd density. This data can be analyzed to detect early signs of disease outbreaks, allowing for prompt public health responses and containment measures.

        \item Behavioral Analysis:
        Authorities can act swiftly and efficiently by recognising odd behaviours or trends that may point to emergencies or health threats by analysing data from monitoring systems.
    \end{itemize}

    \item \textbf{Community Engagement:}
    \begin{itemize}
        \item Transparency:
        In order to interact with the public, cities should be transparent about the data that is gathered, how it is used, and how surveillance technologies work. Policies that are transparent foster confidence and reassure the public about the appropriate application of surveillance technologies.

        \item Education and Participation:
        Community support and cooperation are fostered by informing the public about the advantages of smart city surveillance, such as increased safety and better urban services, and by offering channels for public feedback and involvement in decision-making processes.
    \end{itemize}

    \item \textbf{Crisis Response and Management:}
    \begin{itemize}
        \item Social Media Integration:
        By combining social media analytics with surveillance data, authorities can better assess public opinion during emergencies and respond to community needs.

        \item Drone Surveillance: Deploying drones with surveillance capabilities to quickly analyze the situation from the air during crises like fires, natural catastrophes, or search and rescue missions.
    \end{itemize}
    
\end{itemize}

\section{Model Deployment}

We have created a GUI using the model in our third approach -

\begin{enumerate}
    \item Live camera prediction: Created a GUI using Tkinter which uses the device cam to give live predictions for the labels Figure \ref{fig:p1} \& \ref{fig:p2}. The demo is shown below. The image shown in figure \ref{fig:p3} depicts our setup \href{https://drive.google.com/file/d/1l58YwX20S16Bdf7qPodlvaXMrXW_rkhe/view?usp=sharing}{\textbf{Live demo link}}..

    \begin{figure}[htbp]
        \centering
        \includegraphics[width=0.90\linewidth]{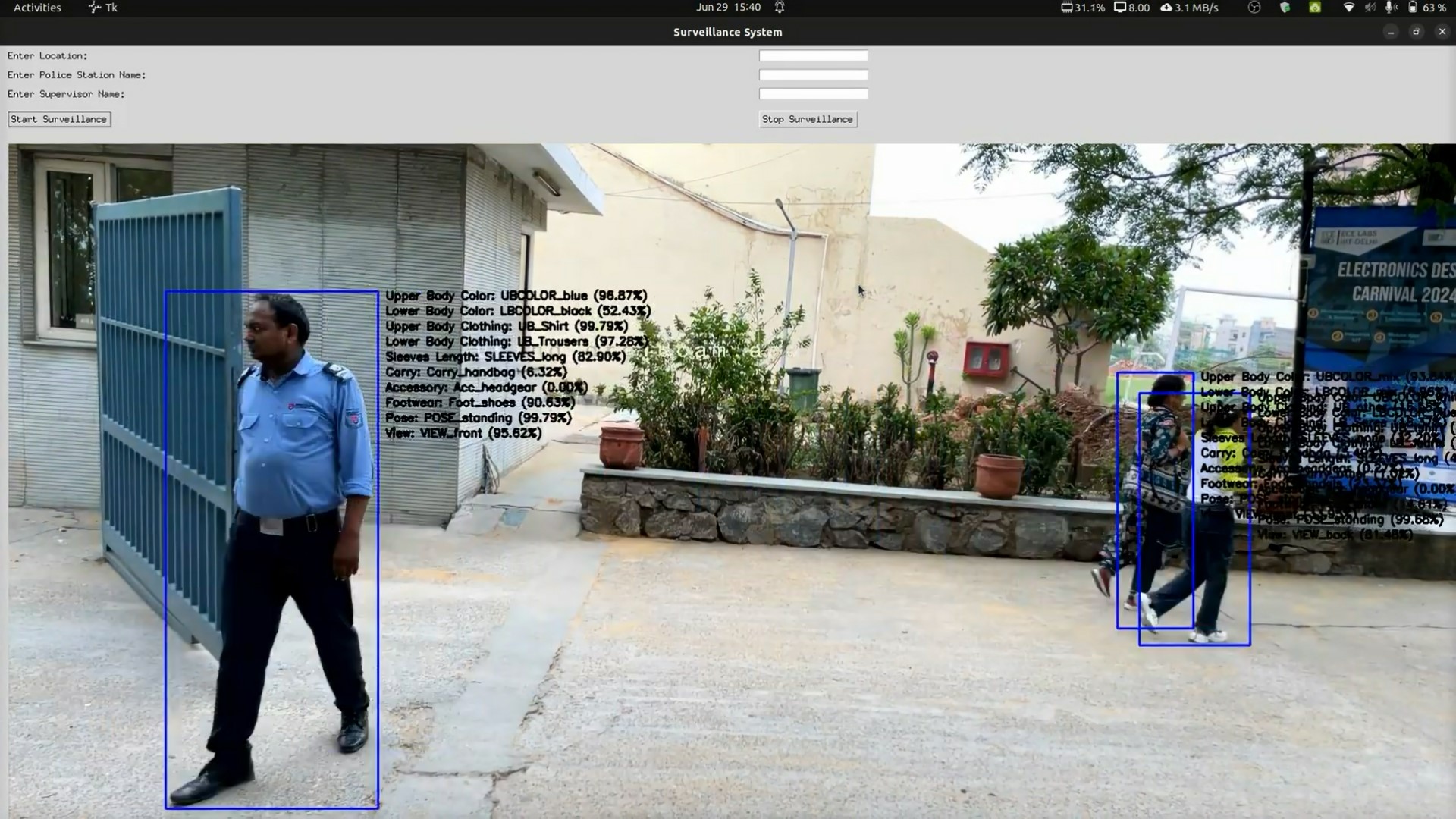}
        \caption{Live Prediction 1}
        \label{fig:p1}
    \end{figure}   

\hspace{2cm}
    \begin{figure}[htbp]
        \centering
        \includegraphics[width=0.90\linewidth]{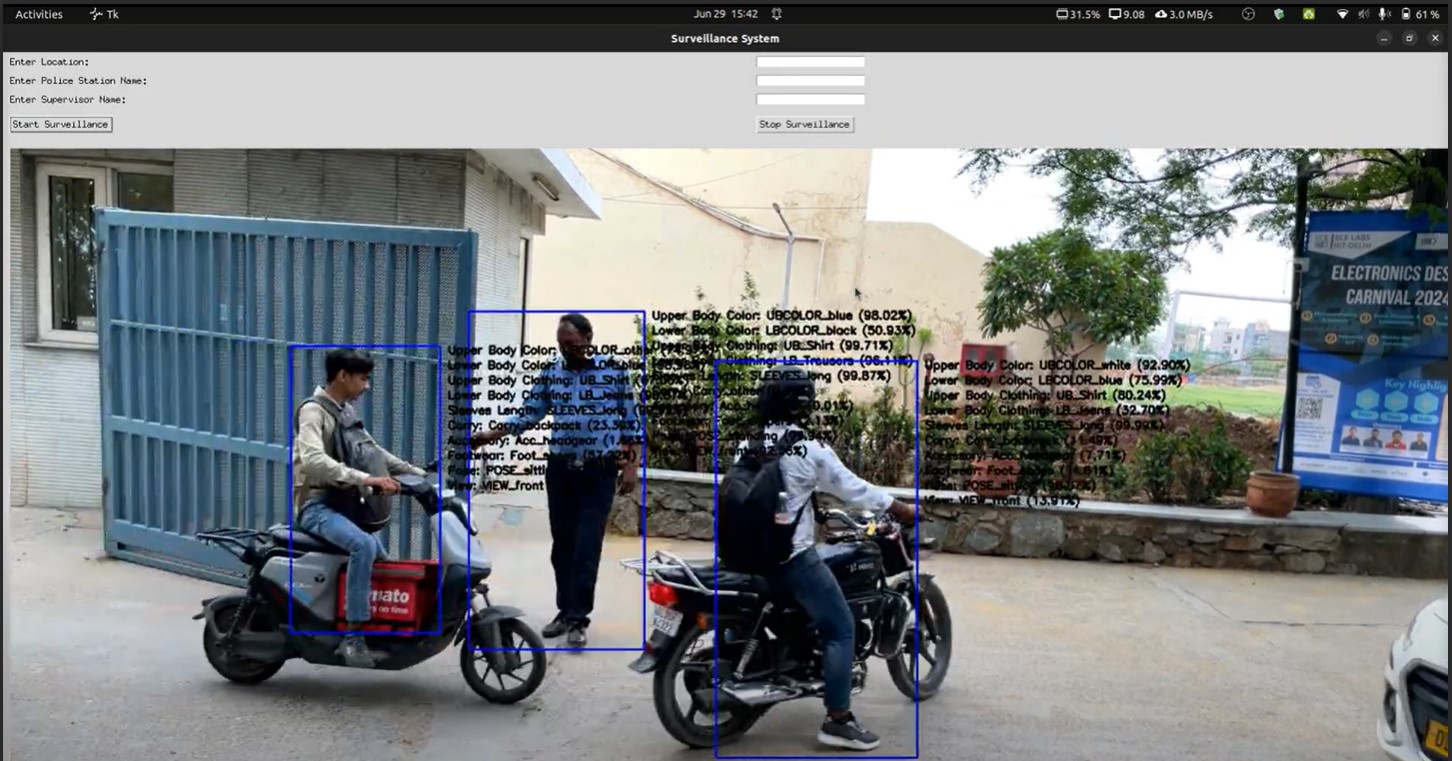}
        \caption{Live Prediction 2}
        \label{fig:p2}
    \end{figure}   

    \begin{figure}[htbp]
        \centering
        \includegraphics[width=0.90\linewidth]{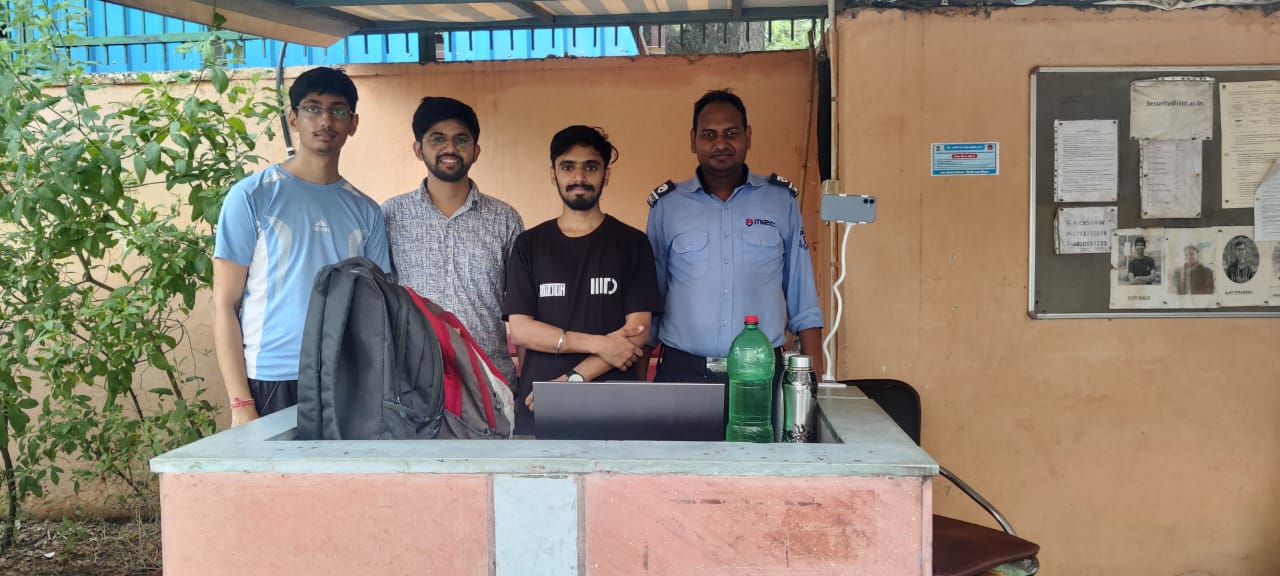}
        \caption{Our setup for live prediction}
        \label{fig:p3}
    \end{figure}   

    \item Static Prediction: We deployed our model on Hugging Face and developed a GUI that enables users to upload images and receive predictions from our model in figure \ref{fig:sp1}.

    \begin{figure}[h!]
        \centering
        \includegraphics[width=0.90\linewidth,height = 17.25cm]{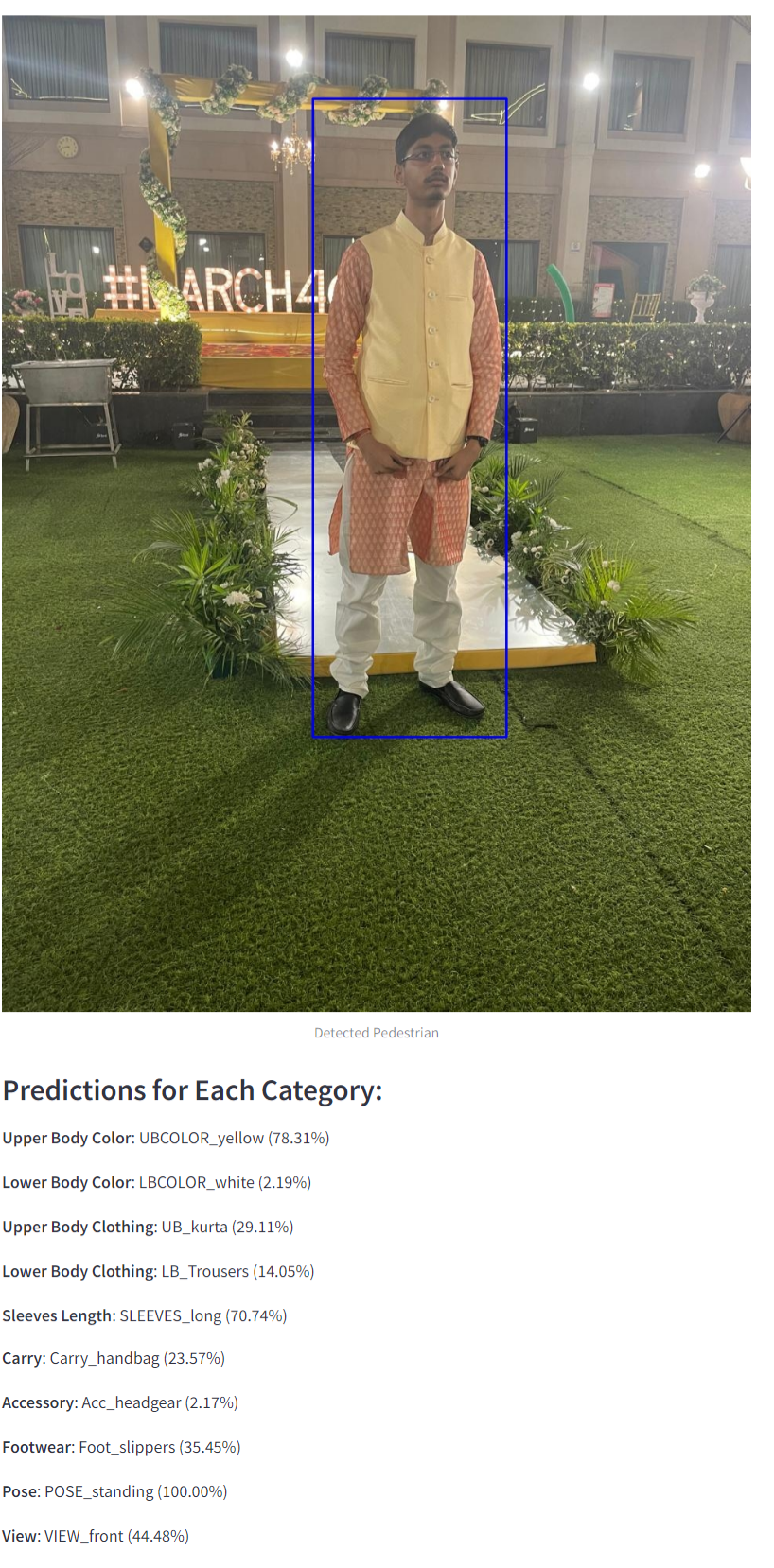}
        \caption{Static Prediction 1}
        \label{fig:sp1}
    \end{figure}

\end{enumerate}

\section{Conclusion}

In this study, we developed a comprehensive pipeline for person attribute recognition, evaluating three distinct approaches: BEiT, SWIN, and FeatClassifier. Our initial experiments with the BEiT model incorporated advanced data augmentation techniques and a novel ScaledBCELoss function to address class imbalance. We then explored the SWIN model, a state-of-the-art architecture renowned for its performance in various vision tasks. However, the FeatClassifier, which integrates a pre-trained ResNet50 backbone with a custom classifier head, emerged as the most effective model. Its superior performance can be attributed to dropout regularization that successfully mitigates overfitting and its pre-training on the RAPv2 dataset, which is specifically comprised of pedestrian images. In contrast, the BEiT and SWIN models were pre-trained on the ImageNet dataset, which contains a diverse range of images.

Our experimental results validate the FeatClassifier approach, demonstrating its strong potential for real-world applications in person attribute recognition. Moving forward, we are excited about further enhancing our pipeline. Future work will involve implementing more sophisticated augmentation strategies, experimenting with various backbone architectures, and extending our model to recognize a wider array of attributes. Additionally, we plan to test our approach on larger and more diverse datasets to further confirm its robustness and scalability. We are committed to advancing this project and look forward to achieving even greater milestones in the field of person attribute recognition.

\bibliographystyle{plain}
\bibliography{ref.bib}

\end{flushright}

\end{document}